\documentclass[10pt,twocolumn,letterpaper]{article}

\usepackage{iccv}
\usepackage{times}
\usepackage{epsfig}
\usepackage{graphicx}
\usepackage{amsmath}
\usepackage{amssymb}

\usepackage{makecell}
\usepackage[numbers]{natbib}
\usepackage{multirow}

\usepackage{amssymb}
\usepackage{pifont}
\newcommand{\cmark}{\ding{51}}%
\newcommand{\xmark}{\ding{55}}%

\usepackage{comment}

\usepackage[pagebackref=true,breaklinks=true,letterpaper=true,colorlinks,bookmarks=false]{hyperref}

\iccvfinalcopy 

\def\iccvPaperID{****} 

\ificcvfinal\fi
\begin{document}

\title{Local Relation Networks for Image Recognition}

\author{Han Hu$^1$ \quad Zheng Zhang$^1$ \quad Zhenda Xie$^{1,2}$ \quad Stephen Lin$^{1}$\\
	$^1$Microsoft Research Asia \hspace{24pt}    $^2$Tsinghua University\\
	{\tt\small \{hanhu, zhez, v-zhxia, stevelin\}@microsoft.com} \\
}

\maketitle

\begin{abstract}

The convolution layer has been the dominant feature extractor in computer vision for years. However, the spatial aggregation in convolution is basically a pattern matching process that applies fixed filters which are inefficient at modeling visual elements with varying spatial distributions. This paper presents a new image feature extractor, called the local relation layer, that adaptively determines aggregation weights based on the compositional relationship of local pixel pairs. With this relational approach, it can composite visual elements into higher-level entities in a more efficient manner that benefits semantic inference. A network built with local relation layers, called the Local Relation Network (LR-Net), is found to provide greater modeling capacity than its counterpart built with regular convolution on large-scale recognition tasks such as ImageNet classification.
\end{abstract}

\section{Introduction}

Humans have a remarkable ability to ``see the infinite world with finite means''~\cite{von1999humboldt,biederman1987recognition}. From perceiving a limited set of low-level visual primitives, they can productively compose unlimited higher-level visual concepts, from which an understanding of a viewed scene can be formed.

In computer vision, this compositional behavior may be approximated by the building of hierarchical representations in a convolutional neural network, where different layers represent different levels of visual elements. At lower layers, basic elements such as edges are extracted. These are combined at middle layers to form object parts, and then finally at higher layers, whole objects are represented~\cite{zeiler2014visualizing}.

\begin{figure}
  \centering
  \includegraphics[width=\linewidth]{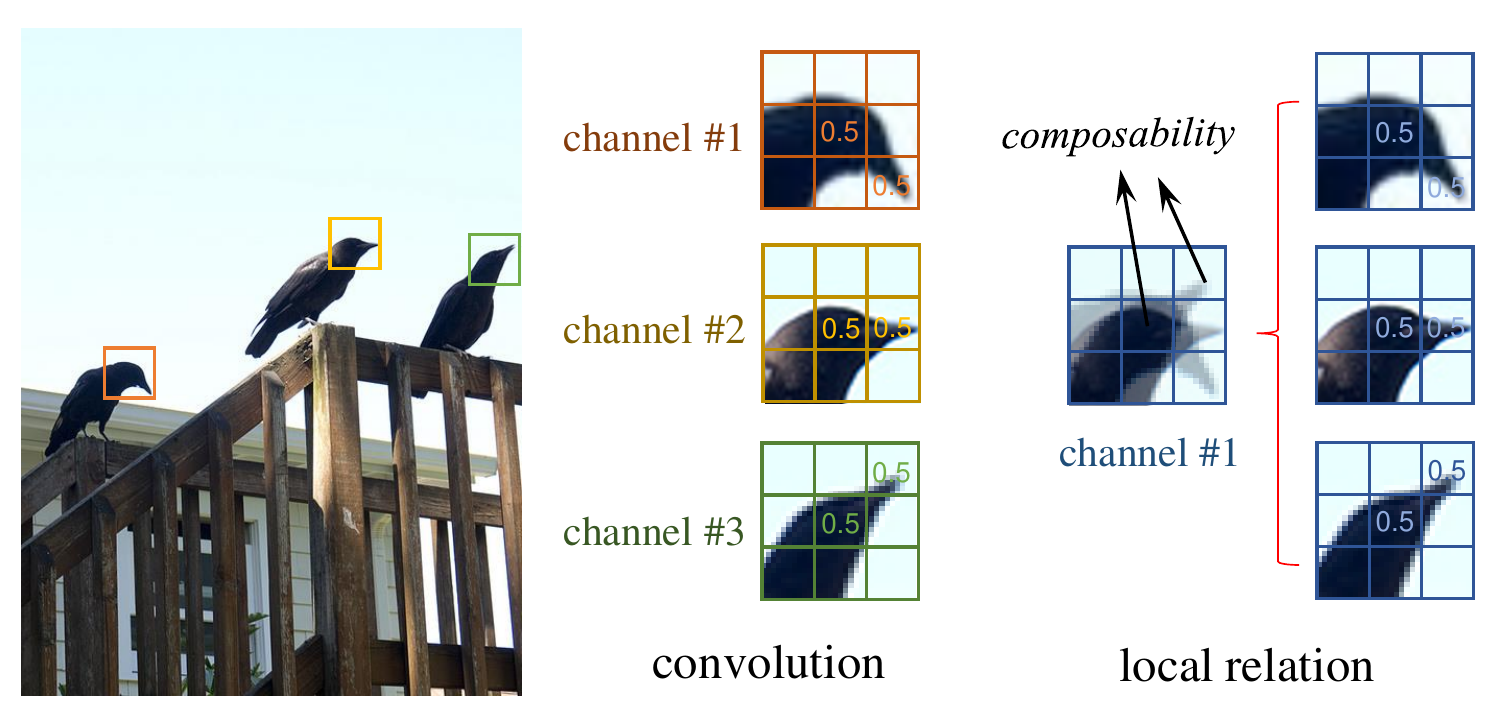}
\caption{Illustration of the 3$\times$3 convolution layer and the 3$\times$3 local relation layer. While 3 channels are required by convolution to represent the spatial variability between bird eye and beak, the local relation layer requires only 1 channel.}
\label{fig.teaser}
\end{figure}

Although a series of convolutional layers can construct a hierarchical representation, its mechanism for composing lower-level elements into higher-level entities can be viewed as highly inefficient in regards to conceptual inference. Rather than recognizing how elements can be meaningfully joined together, convolutional layers act as templates, where input features are spatially aggregated according to convolutional filter weights. For an effective composition of features, suitable filters would need to be learned and applied. This requirement is problematic when trying to infer visual concepts that have significant spatial variability, such as from geometric deformation as illustrated in Fig.~\ref{fig.teaser}, since filter learning could potentially face a combinatorial explosion of different valid compositional possibilities~\cite{sabour2017dynamic,yuille2018deep,marcus2018deep}.

In this paper, we present a new computational network layer, called the local relation layer, in which meaningful compositional structure can be adaptively inferred among visual elements in a local area. In contrast to convolution layers which employ fixed aggregation weights over spatially neighboring input features, our new layer adapts the aggregation weights based on the composability of local pixel pairs. Inspired by recent works on relation modeling~\cite{battaglia2018relational}, composability is determined by the similarity of two pixels' feature projections into a learned embedding space. This embedding may additionally account for geometric priors, which have proven to be useful in visual recognition tasks\footnote{For example, geometric priors are intrinsically encoded in the convolution layer, as its aggregation weights are parameterized on relative positions. This is an important property leading to its success in visual recognition.}. By learning how to adaptively compose pixels in a local area, a more effective and efficient compositional hierarchy can be built. 

Local relation layers can be used as a direct replacement of convolutional layers\footnote{Since $1\times 1$ convolutions do not involve filtering over neighboring pixels, we do not treat them as convolutions in this paper and refer to them as channel transformations~\cite{DBLP:journals/corr/LinCY13}. Nevertheless, in some figures/tables, we use $1\times 1$ to denote a channel transformation layer for notation convenience.} in deep networks, with little added overhead. Using these layers, we have developed a network architecture called Local Relation Network (LR-Net) that follows the practice in ResNet~\cite{he2016deep} of stacking residual blocks to enable optimization of very deep networks. Given the same computation budget, LR-Net with 26 layers and bottleneck residual blocks surpasses the regular 26-layer ResNet by an absolute 3\% in top-1 accuracy on the ImageNet image classification task~\cite{deng2009imagenet}. Improved accuracy is also achieved with basic residual blocks and on deeper networks (50 and 101 layers).

Besides strong image classification performance, we demonstrate several favorable properties of local relation networks. One of them is their greater effectiveness in utilizing large kernel neighborhoods compared to regular convolution networks. While regular ConvNets mainly employ $3 \times 3$ kernels due to saturation at larger sizes, LR-Net is found to benefit from kernels of $7 \times 7$ or even larger. We additionally show that the network is more robust to adversarial attacks, likely due to its compositional power in the spatial domain.

We note that while deep neural networks all form a bottom-up hierarchy of image features, they generally determine feature aggregation weights in a top-down manner. By contrast, our compositional approach computes the weights bottom-up. There exist a few recent methods~\cite{sabour2017dynamic,hinton2018matrix,wang2018non} that also do so, but they are either not applicable to large-scale recognition tasks~\cite{sabour2017dynamic,hinton2018matrix} or act in only a complementary role to regular convolution, rather than as a replacement~\cite{wang2018non}. Moreover, these methods do spatial aggregation over the whole input feature map and do not consider geometric relationships between pixels, while our network demonstrates the importance of \emph{locality} and \emph{geometric priors}.
With this work, it is shown that a bottom-up approach to determining feature aggregation weights can be both practical and effective.

\section{Related Works}

\label{sec.related_works}

\paragraph{Convolution Layers and Extensions}

The convolution layer has existed for several decades~\cite{fukushima1980neocognitron,lecun1989backpropagation}. Its recent popularity started with the impressive performance of AlexNet~\cite{krizhevsky2012imagenet} in classifying objects on ImageNet~\cite{deng2009imagenet}. Since then, the convolution layer has been almost exclusively used in extracting basic visual features. 

Extensions to the regular convolution layer have been proposed. In one direction, a better accuracy-efficiency tradeoff is obtained by limiting the scope of aggregated input channels. Representative works include group convolution~\cite{krizhevsky2012imagenet,xie2017aggregated} and depthwise convolution~\cite{chollet2017xception,howard2017mobilenets}.
Another direction is to modify the spatial scope for aggregation. This has been done to enlarge the receptive field, such as through atrous/dilated convolution~\cite{chen2018deeplab,yu2015multi}, and to enhance the ability to model geometric deformation, via active~\cite{jeon2017active} and deformable convolution~\cite{dai2017deformable,zhu2018deformable}.

Some works relax the requirement of sharing aggregation weights/scopes across positions. A straightforward approach is taken with the locally connected layer~\cite{taigman2014deepface}, which learns independent aggregation weights for different positions. Its application is limited due to the loss of important properties from regular convolution, including translation invariance and knowledge transfer from one position to others. In other works along this direction, convolution layers are proposed which generate position-adaptive aggregation weights~\cite{jia2016dynamic} or an adaptive aggregation scope~\cite{dai2017deformable,zhu2018deformable}.

We note that regular convolution and the above extensions all operate in a top-down manner, determining their convolution behavior based on image appearance or spatial positions within a receptive field. In contrast, the proposed layer determines aggregation weights in a bottom-up fashion based on composability of local pixel pairs, which we believe provides a more efficient encoding of spatial composition in the visual world. At the same time, the proposed layer follows and adapts several favorable design principles from these convolution variants, such as \emph{locality}, use of \emph{geometric priors}, and weight/meta-weight sharing across positions, which have been found to be crucial in effectively extracting visual features.

\paragraph{Capsule Networks}

To address some shortcomings of convolution layers, there have been recent works that determine the aggregation weights in a bottom-up manner based on the composability of pixel pairs. A representative work is Capsule Networks~\cite{sabour2017dynamic,hinton2018matrix}, in which composability is computed by an iterative routing process. In each routing step, the aggregation weights are enlarged if the vectors before and after aggregation are close to each other, and they are reduced otherwise. This self-strengthening process in capsule networks is similar to the process of a \emph{filtering bubble}, a popular phenomenon in social networks where the connection between agents with the same interests becomes stronger, while the connections become weaker when interests are dissimilar.

Although the routing method is inspiring, the computation is not well aligned with current learning infrastructure such as back-propagation and multi-layer networks. In contrast, the composability of pixel pairs in the local relation layer is computed by the similarity of pixel pairs in an embedding space with learnt embedding parameters, which is more friendly to current learning infrastructure. The local relation layer is also differentiated from capsule networks by its aggregation computation process, including its spatial scope (\emph{local} vs.~\emph{global}) and geometric priors (\emph{with vs.~without}). With these differences, local relation networks are significantly more practical than existing methods based on bottom-up aggregation.

\begin{table*}[t]
	\setlength{\tabcolsep}{4pt}
	\centering
    \caption{A summary of basic image feature extractors. The ``parameterization'' column indicates the model weights to be learnt. The symbols $\omega, \{\theta\}, \Omega$ denote aggregation weights, weights of meta networks, and spatial sampling points, respectively. ``share'' indicates whether the parameterized weights are shared across position. The aggregation scope is given over both the channel and spatial domains. The ``aggregation weight'' column covers three aspects: how aggregation weights are computed from parameterized weights (``computation'' sub-column); inclusion of geometric priors (``geo.'' sub-column); type of computation (``type'' sub-column).}
    \begin{tabular}{c|l|c|c|c|c|c|c|c}
		\Xhline{2\arrayrulewidth}
	    \multicolumn{1}{c}{} & \multirow{2}{*}{method} & \multicolumn{2}{c|}{parameterization} & \multicolumn{2}{c|}{aggregation scope} & \multicolumn{3}{c}{aggregation weight} \\
	    \cline{3-9}
	    \multicolumn{2}{c|}{} & param. & share & \makecell{channel\\(in/out/share)} & spatial & computation & geo. & type \\
        \Xhline{2\arrayrulewidth}
	    \multirow{8}{*}{conv.} & regular & $\omega$ & \checkmark & all/one/no & local & $\omega$ & \checkmark & top-down\\
	    \cline{2-9}
	    & group~\cite{krizhevsky2012imagenet,xie2017aggregated} & $\omega$ & \checkmark & \makecell{group/one/no} & local & $\omega$ &\checkmark & top-down \\
	    \cline{2-9}
	    & depthwise~\cite{chollet2017xception,howard2017mobilenets} & $\omega$ & \checkmark & \makecell{one/one/no} & local & $\omega$ &\checkmark & top-down \\
	    \cline{2-9}
	    & dilated~\cite{chen2018deeplab,yu2015multi} & $\omega$ & \checkmark & all/one/no & \makecell{atrous} & $\omega$ & \checkmark & top-down \\
	    \cline{2-9}
	    & active~\cite{jeon2017active} & $\omega$, $\Omega$ & \checkmark & all/one/no & $\Omega$ & $\omega$ & \checkmark &top-down \\
	    \cline{2-9}
	    & local connected~\cite{taigman2014deepface} & $\omega$ & \xmark & all/one/no & local & $\omega$ & \checkmark &top-down \\
	    \cline{2-9}
	    & dynamic filters~\cite{jia2016dynamic} & $\theta$ & \checkmark & all/one/no & local & $f_{\theta}(\mathbf{x}_{\mathbf{p}'})$ & \checkmark &top-down \\
	    \cline{2-9}
	    & deformable~\cite{dai2017deformable,zhu2018deformable} & $\omega$, $\theta$ & \checkmark & all/one/no & $\Omega(\theta)$ & $\omega$ & \checkmark &top-down \\
	    \Xhline{2\arrayrulewidth}
	    \multicolumn{1}{c}{} & non-local~\cite{wang2018non} & $\theta_k, \theta_q$ & \checkmark & \makecell{one/one/all} & full & $\Phi(f_{\theta_{q}}(\mathbf{x}_{\mathbf{p}'}), f_{\theta_k}(\mathbf{x}_{\mathbf{p}}))$ & \xmark & bottom-up \\
	    \hline
	    \multicolumn{1}{c}{} & capsule~\cite{sabour2017dynamic,hinton2018matrix} & $\theta$ & \xmark & \makecell{one/one/group} & full & $\text{route}(\mathbf{y}_{\mathbf{p}'}, f_{\theta}(\mathbf{x}_{\mathbf{p}}))$ & \xmark & bottom-up \\
		\Xhline{2\arrayrulewidth}
		\multicolumn{1}{c}{} & local relation (our) &  $\theta_k, \theta_q, \theta_g$ & \checkmark & \makecell{one/one/group} & local & \makecell{$\text{softmax}_{\Omega}(\Phi(f_{\theta_{q}}(\mathbf{x}_{\mathbf{p}'}),$ \\ $f_{\theta_k}(\mathbf{x}_{\mathbf{p}})) + f_{\theta_g}(\mathbf{p}-\mathbf{p}'))$} & \checkmark & bottom-up \\
		\Xhline{2\arrayrulewidth}
	\end{tabular}
	\label{table:common_layers}
\end{table*}

\paragraph{Self-Attention / Graph Networks}

The proposed local relation layer is also related to self-attention models~\cite{vaswani2017attention} used in natural language processing, and to graph networks applied on non-grid data~\cite{bronstein2017geometric}. These works share a basic structure similar to general relation modeling~\cite{battaglia2018relational}, which naturally introduces compositionality in the networks.

Due mainly to their powerful composition modeling ability, these methods have become the dominant approaches in their respective fields. However, in computer vision, there are few works involving such compositionality in their network architecture~\cite{hu2018relation, wang2018non}. In~\cite{hu2018relation}, relationships between object proposals are modeled, which leads to improved accuracy as well as the first fully end-to-end object detector. The relation modeling in that work is applied to non-grid data. In~\cite{wang2018non}, relationships are modeled between pixels, as in our work. However, the goal is different. While~\cite{wang2018non} extracts long-range context as complementary to the convolution layer, we pursue a basic image feature extractor with more representation power for spatial composition  than the convolution layer.

In this sense, our work bridges the general philosophy of introducing compositionality into representation, which has proven effective in processing sequential and non-grid data, and applicability as a basic feature extractor for computer vision. Such a goal is non-trivial and requires adaptations from both sides.

\section{A General Formulation}

In this section, we describe a general formulation for basic image feature extractors, based on which the proposed local relation layer will be presented. Denote the input and output of a layer by $\mathbf{x} \in \mathbb{R}^{C \times H \times W}$ and $\mathbf{y} \in \mathbb{R}^{C' \times H' \times W'}$, with $C, C'$ being the channels of input/output features and $H, W, H', W'$ the intput/output spatial resolution. Existing basic image extraction layers generally produce the output feature by a weighted aggregation of input features,
\begin{equation}
    \label{eq.general_formulation}
\mathbf{y} (c', \mathbf{p}') = \sum_{c \in \Omega_{c'}, \mathbf{p} \in \Omega_{\mathbf{p}'}} \mathbf{\omega} (c', c, \mathbf{p}', \mathbf{p}) \cdot \mathbf{x} (c, \mathbf{p}),
\end{equation}
where $c, c'$ and $\mathbf{p}=(h, w),\mathbf{p}'=(h', w')$ index the input and output channels and feature map positions, respectively; $\Omega_{c'}$ and $\Omega_{\mathbf{p}'}$ denote the scope for channel and spatial aggregation of input features in producing the output feature value at channel $c'$ and position $\mathbf{p}'$, respectively; $\omega(c', c, \mathbf{p}', \mathbf{p})$ denotes the aggregation weight from $c, \mathbf{p}$ to $c', \mathbf{p}'$. Existing basic image feature extraction layers are differentiated mainly by three aspects: parameterization method, aggregation scope, and aggregation weights. 

\paragraph{Parameterization method} defines the model weights to be learnt. The most common parameterization method is to directly learn the aggregation weights $\omega$~\cite{lecun1989backpropagation}. There are also some methods that learn a meta network $\{\theta\}$ on input features to generate adaptive aggregation weights~\cite{jia2016dynamic} or an adaptive aggregation scope across spatial positions~\cite{dai2017deformable}, or learn a fixed prior about spatial aggregation scope ($\Omega$)~\cite{jeon2017active}. In general, the parameterization is shared across spatial position to enable translation invariance.

\paragraph{Aggregation scope} defines the range of channels and spatial positions involved in aggregation computation. For channel scope, regular convolution includes all input channels in computing each channel output. For greater efficiency, some methods consider only one or a group of input channels in producing one channel of the output feature~\cite{krizhevsky2012imagenet,chollet2017xception}. Recently, there have been methods where multiple or all output channels share the same aggregation weights~\cite{wang2018non,sabour2017dynamic}. For spatial scope, most methods constrain the aggregation computation in a local area. Restricting aggregation to a local area can not only significantly reduce computation, but also help introduce an information bottleneck that facilitates learning of visual patterns. Nevertheless, recent non-convolution methods~\cite{wang2018non,sabour2017dynamic} mostly adopt a full spatial scope for aggregation computation.

\paragraph{Aggregation weights} are typically learned as network parameters or are computed from them. Almost all variants of convolution obtain their aggregation weights in a \emph{top-down} manner, where they are either fixed across positions or determined by a meta network on the input features at the position. There are also some non-convolution methods~\cite{wang2018non,sabour2017dynamic} that compute the aggregation weights in a \emph{bottom-up} fashion, with the weights determined by the composability of a pixel pair. In contrast to convolution variants whose aggregation weights depend heavily on geometric priors, such priors are seldom used in recent non-convolution methods.

Table~\ref{table:common_layers} presents a summary of existing basic image feature extractors.

\section{Local Relation Layer}
\label{sec.local_relation_layer}

In this section, we introduce the local relation layer. Expressed within the general formulation of Eqn.~(\ref{eq.general_formulation}), its aggregation weights are defined as\footnote{Since one output channel strictly uses one input channel in aggregation computation, we omit the $c, c'$ for notational convenience.}
\begin{equation}
\label{eq.local_relation_aggregation_weight}
\mathbf{\omega} (\mathbf{p}', \mathbf{p}) = 
    \text{softmax}(\Phi(f_{\theta_{q}}(\mathbf{x}_{\mathbf{p}'}), f_{\theta_k}(\mathbf{x}_{\mathbf{p}})) + f_{\theta_g}(\mathbf{p}-\mathbf{p}')),
\end{equation}
where the term $\Phi(f_{\theta_{q}}(\mathbf{x}_{\mathbf{p}'}), f_{\theta_k}(\mathbf{x}_{\mathbf{p}}))$ is a measure of composability between the target pixel $\mathbf{p}'$ and a pixel $\mathbf{p}$ within its position scope, based on their appearance after transformations $f_{\theta_{q}}$ and $f_{\theta_k}$, following recent works on relation modeling~\cite{battaglia2018relational}. The term $f_{\theta_g}(\mathbf{p}-\mathbf{p}')$ defines the composability of a pixel pair $(\mathbf{p}, \mathbf{p}')$ based on a geometric prior. The geometric term adopts the relative position as input and is translationally invariant.

This new layer belongs to class of bottom-up methods, as indicated in Table~\ref{table:common_layers}, as it determines composability based on the properties of the two visual elements. In the following, we present its design and discuss its differences from existing bottom-up methods. These differences lead to significant higher accuracy on image recognition benchmarks. Its performance also is comparable to or surpasses state-of-the-art top-down convolution methods.

\paragraph{Locality} The bottom-up methods typically aggregate input features from over the full image. In contrast, the local relation layer limits the aggregation computation to a local area, e.g., a $7\times7$ neighborhood. We find that constraining the aggregation scope to a local neighborhood is crucial for feature learning in visual recognition (see Table~\ref{table:locality}).

Compared with the convolution variants which also constrain the aggregation computation to a spatial neighborhood, the local relation layer proves more effective in utilizing larger kernels. While convolution variants usually exhibit performance saturation with neighborhoods larger than $3\times 3$, the local relation layer yields steady improvements in accuracy when increasing the neighborhood size from $3 \times 3$ to $7\times 7$ (see Table~\ref{table:locality}). This difference may be due to the representation power of convolution layer being bottlenecked by the number of fixed filters, hence there is no benefit from a larger kernel size. In contrast, the local relation layer composes local pixel pairs in a flexible bottom-up manner that allows it to effectively model visual patterns of increasing size and complexity. We use a $7\times7$ kernel size by default.

\paragraph{Appearance composability}

We follow a general approach for relation modeling~\cite{battaglia2018relational} to compute appearance composability $\Phi(f_{\theta_{q}}(\mathbf{x}_{\mathbf{p}'}), f_{\theta_k}(\mathbf{x}_{\mathbf{p}}))$, where $\mathbf{x}_{\mathbf{p}'}$ and $\mathbf{x}_{\mathbf{p}}$ are projected to a \emph{query} (by a channel transformation layer $f_{\theta_{q}}$) and \emph{key} (by a channel transformation layer $f_{\theta_{k}}$) embedding space, respectively. While in previous works the \emph{query} and \emph{key} are vectors, in the local relation layer, we use scalars to represent them so that the computation and representation are lightweight. We find that scalars work also well and have better speed-accuracy trade-off compared to vectors (see Table~\ref{table:key_dim}).

We consider the following instantiations of function $\Phi$,  which we later show to work similarly well (see Table~\ref{table:app_term}):

~~~~a) squared difference:
\begin{equation}
\label{eq.square_diff}
    \Phi(q_{\mathbf{p}'}, k_{\mathbf{p}}) = -( q_{\mathbf{p}'} - k_{\mathbf{p}})^2.
\end{equation}

~~~~b) absolute difference:
\begin{equation}
\label{eq.abs_diff}
    \Phi(q_{\mathbf{p}'}, k_{\mathbf{p}}) = -| q_{\mathbf{p}'} - k_{\mathbf{p}}|,
\end{equation}

~~~~c) multiplication:
\begin{equation}
\label{eq.multiplication}
    \Phi(q_{\mathbf{p}'}, k_{\mathbf{p}}) = q_{\mathbf{p}'} \cdot k_{\mathbf{p}},
\end{equation}

We use Eqn.~(\ref{eq.square_diff}) by default.

\paragraph{Geometric priors} Another important aspect differentiating the local relation layer from other bottom-up methods is the inclusion of geometric priors.

\begin{figure}
  \centering
  \includegraphics[width=\linewidth]{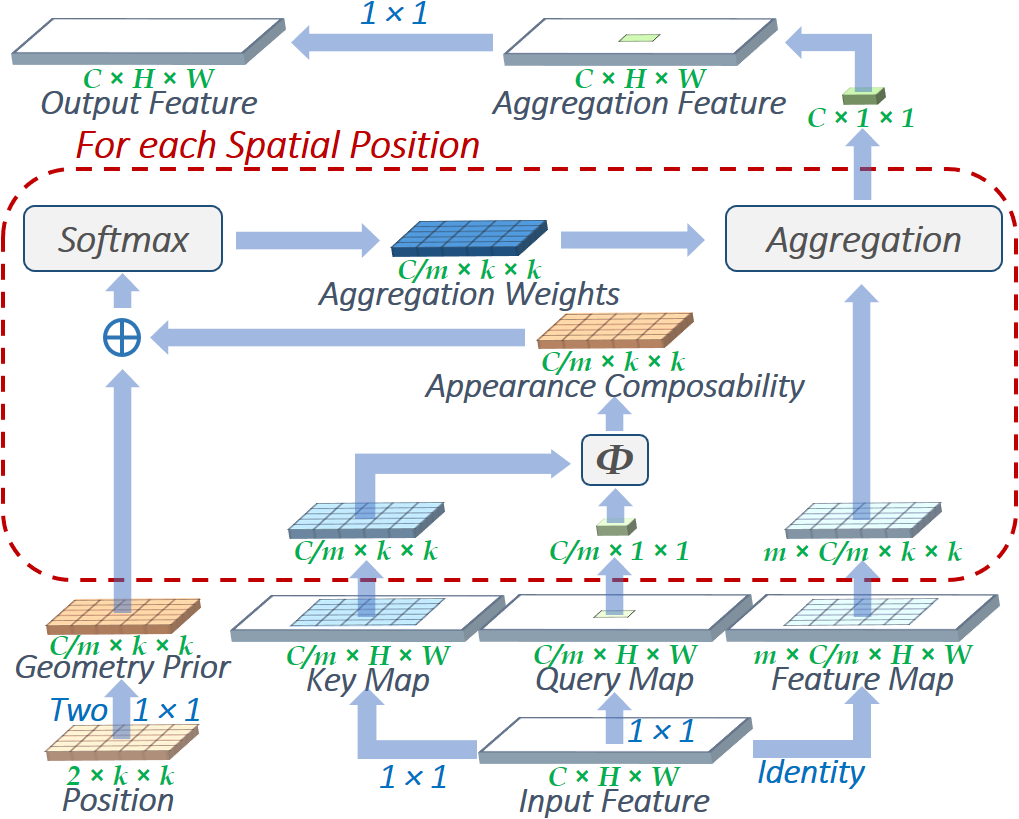}
\caption{The local relation layer.}
\label{fig.local_relation_layer}
\end{figure}

The geometric prior is encoded by a small network on the relative position of $\mathbf{p}$ to $\mathbf{p}'$. The small network consists of two channel transformation layers, with a ReLU activation in between. We find that using a small network to compute the geometric prior values is better than directly learning the values, especially when the neighborhood size is large (see Table~\ref{table:locality}). This is possibly because a small network on relative position treats relative positions as vectors in metric space, while the direct method treats different relative positions as independent identities.

Note that the inference process with using a small network is the same as that of directly learning the geometric priors. In fact, during inference, the fixed learnt weights $\theta_g$ will induce fixed geometric prior values $f_{\theta_g}(\Delta \mathbf{p})$ for a relative position $\Delta \mathbf{p}$. We use these fixed geometric prior values instead of the original model weights $\theta_g$ for more convenient inference.

\paragraph{Weight normalization} We use SoftMax normalization over the spatial scope $\Omega$ to compute the final aggregation weights. Such normalization is found to be crucial in balancing the contributions of the appearance composability and geometric prior terms (see Table~\ref{table:app_term}).

\paragraph{Channel sharing} Following~\cite{sabour2017dynamic}, the local relation layer uses \emph{channel sharing} in aggregation computation, where multiple channels share the same aggregation weights.  \emph{Channel sharing} can generally reduce model size and facilitate GPU memory scheduling for efficient implementation. We observe no accuracy drop with up to 8 channels (default) sharing the same aggregation (see Table~\ref{table:group_num}), while achieving more than 3$\times$ actual speed-up than that of 1 channel per aggregation in our CUDA kernel implementation.

\paragraph{Complexity and implementation}

The local relation layer is summarized in Figure~\ref{fig.local_relation_layer}. Given an $H\times W$ input feature map, $k\times k$ spatial neighborhood, $C$ channels, and $m$ channels per aggregation computation, the total computational complexity (in FLOPs) of a local relation layer with stride $s$ is
\begin{equation}
    \mathcal{C} = \mathcal{O}\left((\frac{1+s^2}{m}+1)C(C+k^2)\frac{HW}{s^2} \right).
\end{equation}

In our experiments, a naive implementation by a CUDA kernel is used, which is several times slower than regular convolution with the same FLOPs\footnote{The LR-Net-26 network introduced in Section~\ref{sec.relation_networks} is about 3$\times$ slower than that of a regular ResNet-26 model on a Titan Xp GPU.}. Note that convolution has a highly optimized implementations with careful memory scheduling. Optimization of memory scheduling for the local relation layer will be a focus of our future work.

\section{Local Relation Networks}

\label{sec.relation_networks}

Local relation layers can be used to replace spatial convolution layers in deep neural networks. In this section, we describe layer replacement in the ResNet architecture~\cite{he2016deep}, where residual blocks with the same topology are stacked.

\begin{figure}
  \centering
  \includegraphics[width=\linewidth]{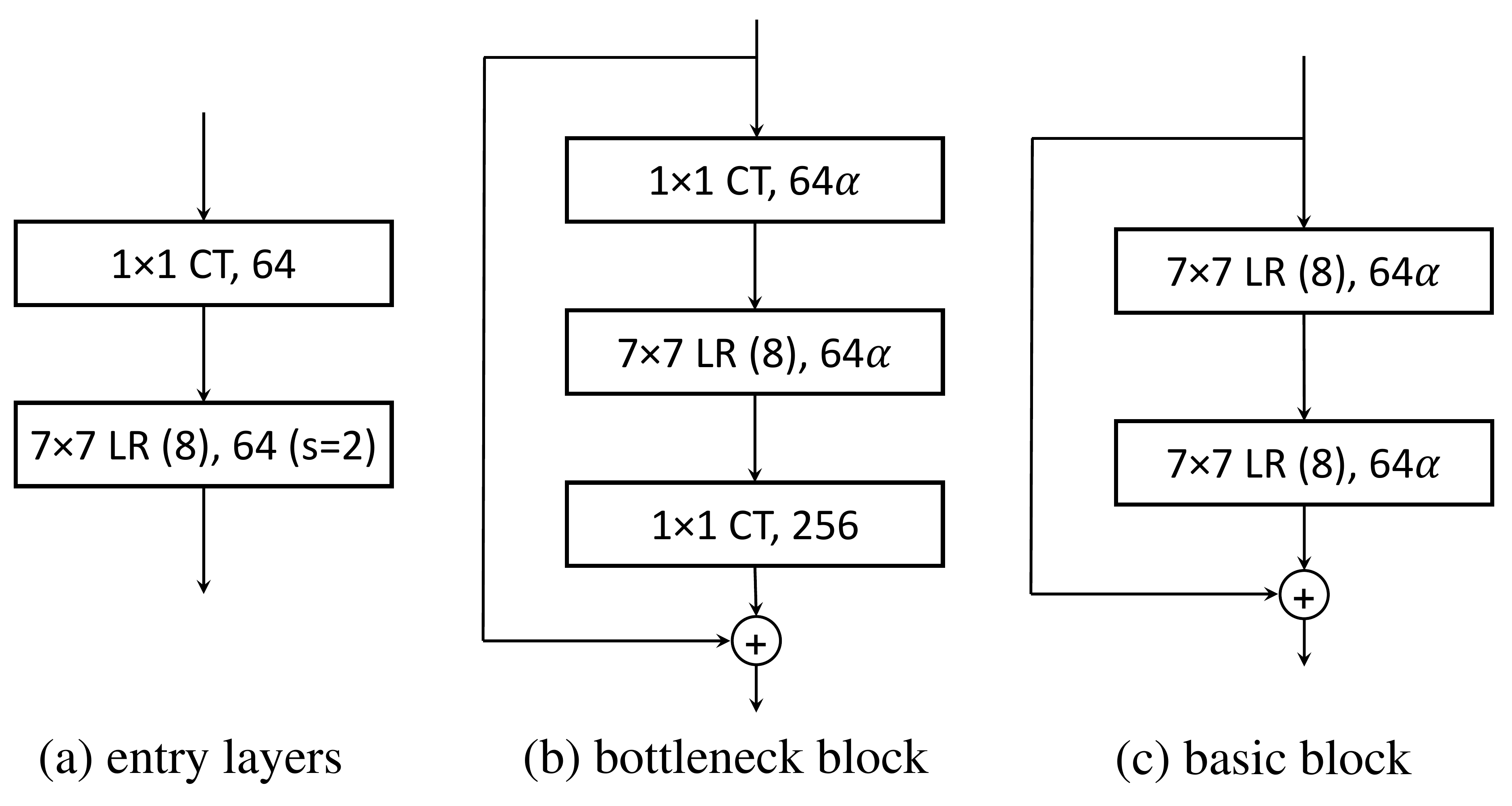}
\caption{Illustration of replacing the first $7\times 7$ convolution layer (a) and the bottleneck/basic residual blocks (b)(c) in the ResNet architecture. ``CT'' denotes the channel transformation layer and ``LR'' denotes the local relation layer. ``7$\times$7 (8), 64'' represents kernel size of 7$\times$7, channel sharing of $m=8$ and output channel of 64. ``$s=2$" represents a stride of 2. All layers are followed by a batch normalization layer and a ReLU activation layer.}
\label{fig.replace_blocks}
\end{figure}

Figure~\ref{fig.replace_blocks} illustrates the replacement of the $3\times 3$ convolution layer in the bottleneck/basic residual blocks and the first $7\times 7$ convolution layer in ResNet. For residual blocks, we keep the FLOPs the same by adapting the expansion ratio ($\alpha$) of the layer to be replaced. For the first $7\times 7$ convolution layer, we transform the $3\times H \times W$ input to a feature map of $64 \times H \times W$ by a channel transformation layer, and follow this with a $7\times 7$ local relation layer. The replacement of the $7\times 7$ convolution layer consumes similar FLOPs and has comparable accuracy on ImageNet recognition. In the experiments, we will mainly ablate the effects of replacing $3\times 3$ convolution layers in residual blocks.

After replacing all convolution layers in ResNet, we obtain a network which we call the Local Relation Network (LR-Net). Table~\ref{table.lr_net50} shows a comparison of ResNet-50 and LR-Net-50 (with default hyper-parameters of $7\times 7$ kernel size and $m=8$ channels per aggregation). LR-Net-50 uses similar FLOPs but has a slightly smaller model size because of its \emph{channel sharing} in aggregation.

\newcommand{\blockb}[3]{\multirow{3}{*}{
\(\left[
\begin{array}{l}
\text{1$\times$1, #2}\\
[-.1em] \text{3$\times$3 conv, #2}\\
[-.1em] \text{1$\times$1, #1}
\end{array}\right]\)$\times$#3}
}

\newcommand{\blockx}[3]{\multirow{3}{*}{
\(\left[
\begin{array}{l}
\text{1$\times$1, #2}\\
[-.1em] \textbf{7$\times$7 LR, #2}\\
[-.1em] \text{1$\times$1, #1}\\
\end{array}\right]\)$\times$#3}
}

\newcolumntype{x}[1]{>\centering p{#1pt}}
\newcommand{\ft}[1]{\fontsize{#1pt}{1em}\selectfont}
\renewcommand\arraystretch{1.25}
\setlength{\tabcolsep}{1.2pt}
\begin{table}[t]
\begin{center}
\footnotesize
\begin{tabular}{c|c|x{80}|c}
\hline
 stage & output & ResNet-50 & \textbf{LR-Net-50 (7$\times$7, $m$=8)} \\
\hline
res1 & \ft{7} 112$\times$112 & 7$\times$7 conv, 64, stride 2 & \makecell{\textbf{1$\times$1, 64} \\ \textbf{7$\times$7 LR, 64, stride 2}} \\
\hline
\multirow{4}{*}{res2} & \multirow{4}{*}{\ft{7} 56$\times$56} & 3$\times$3 max pool, stride 2 & 3$\times$3 max pool, stride 2 \\\cline{3-4}
  &  &  \blockb{256}{64}{3} & \blockx{256}{100}{3}\\
  &  &  & \\
  &  &  & \\
\hline
\multirow{3}{*}{res3} &  \multirow{3}{*}{\ft{7} 28$\times$28}
  & \blockb{512}{128}{4} &  \blockx{512}{200}{4}\\
  &  &  & \\
  &  &  & \\
\hline
\multirow{3}{*}{res4} & \multirow{3}{*}{\ft{7} 14$\times$14}
  & \blockb{1024}{256}{6} & \blockx{1024}{400}{6}\\
  &  &  & \\
  &  &  & \\
\hline
\multirow{3}{*}{res5} & \multirow{3}{*}{\ft{7} 7$\times$7}
& \blockb{2048}{512}{3} & \blockx{2048}{800}{3}\\
  &  &  & \\
  &  &  & \\
\hline
& \multirow{2}{*}{\ft{7} 1$\times$1} & global average pool & global average pool \\
 & & 1000-d fc, softmax & 1000-d fc, softmax \\
\hline
\multicolumn{2}{c|}{\small \# params} & \small \textbf{25.5}$\times$$10^6$  & \small \textbf{23.3}$\times$$10^6$ \\
\hline
\multicolumn{2}{c|}{\small FLOPs} & \small \textbf{4.3}$\times$$10^9$  & \small \textbf{4.3}$\times$$10^9$ \\
\hline
\end{tabular}
\end{center}
\caption{(\textbf{Left}) ResNet-50. (\textbf{Right}) LR-Net-50 with $7\times 7$ kernel size and $m=8$ channels per aggregation computation. Inside the brackets are the shape of a residual block, and outside the brackets is the number of stacked blocks in a stage. \emph{LR-Net-50 requires similar FLOPs as ResNet-50 and a slightly smaller number of parameters.}
}
\label{table.lr_net50}
\vspace{-.5em}
\end{table}

\section{Experiments}

We perform an ablation study on the ImageNet-1K image classification task. To facilitate the study given limited GPU resources, we conduct the study using LR-Net-26, which is a 26 layer local relation network adapted from ResNet-26. The networks have 8 bottleneck residual blocks, with $\{2,2,2,2\}$ blocks for \emph{res2}, \emph{res3}, \emph{res4}, \emph{res5}, respectively. We also report results on networks stacked by basic residual blocks (LR-Net-18) and with larger depth of layers (LR-Net-50, LR-Net-101). The robustness of LR-Nets to adversarial attacks is examined as well.

Our experimental settings and hyper-parameters mostly follow~\cite{xie2017aggregated}. Please see the appendix for details.

\begin{table*}[t]
	\setlength{\tabcolsep}{8pt}
	\centering
    \caption{Recognition performance of different architectures with varying spatial aggregation scope and different geometric prior terms on ImageNet classification. Top-1 and top-5 accuracy (\%) is reported. ``NG'' denotes local relation networks without the geometric prior term. ``G*'' represents the method that directly learns the geometric prior values as described in Section~\ref{sec.local_relation_layer}. For fair comparison, we set all the architectures to have similar FLOPs with the regular ResNet-26 model, by adapting their bottleneck ratio $\alpha$. For ResNet-(DW)-26 networks, we omit the ``full image'' column due to implementation difficulty. }
    \begin{tabular}{c|c|c|c|c|c|c|c|c|c|c|c}
		\Xhline{2\arrayrulewidth}
	    \multirow{3}{*}{\makecell{network}} & \multirow{3}{*}{\makecell{geo.\\prior}} & \multicolumn{10}{c}{aggregation spatial scope} \\
	    \cline{3-12}
	    & & \multicolumn{2}{c|}{$3\times3$} & \multicolumn{2}{c|}{$5\times 5$} & \multicolumn{2}{c|}{$7\times 7$} & \multicolumn{2}{c|}{$9\times 9$} & \multicolumn{2}{c}{full image} \\
	    \cline{3-12}
	    & & top-1 & top-5 & top-1 & top-5 & top-1 & top-5 & top-1 & top-5 & top-1 & top-5 \\
		\hline 
        ResNet-26 & \cmark & 72.8 & 91.0 & 73.0 & 91.1 & 72.3 & 90.7 & 71.4 & 90.3 & - & -\\
        ResNet-DW-26 & \cmark & 73.7 & 91.5 & 73.9 & 91.6 & 73.8 & 91.6 & 73.8 & 91.6 & - & - \\
        \hline 
        LR-Net-26 (NG) & \xmark & 70.8 & 89.8 & 71.5 & 90.1 & 71.9 & 90.4 & 70.2 & 89.3 & 50.7 & 74.7 \\
        LR-Net-26 (G*) & \cmark & 73.2 & 91.1 & 74.1 & 91.7 & 73.6 & 91.2 & 72.3 & 90.7 & 60.3 & 82.1 \\
        LR-Net-26 & \cmark & 73.6 & 91.6 & 74.9 & 92.3 & \textbf{75.7} & \textbf{92.6} & 75.4 & 92.4 & 68.4 & 88.0 \\
		\Xhline{2\arrayrulewidth}
	\end{tabular}
	\label{table:locality}
	\vspace{-.5em}
\end{table*}

\subsection{Ablation Study}

\paragraph{Impact of spatial scope} Table~\ref{table:locality} presents the impact of varying aggregation spatial scope for the proposed local relation networks, as well as the regular ResNet-26 network and its variant, ResNet-DW-26~\cite{ma2018shufflenet}, where the regular convolution layer is replaced by depthwise convolution. We have the following observations.

\begin{table}[t]
	\setlength{\tabcolsep}{6pt}
	\centering
    \caption{ Ablation on \emph{query}/\emph{key} dimension (top-1 acc \%).}
    \begin{tabular}{c|c|c|c|c|c}
		\Xhline{2\arrayrulewidth}
	    \makecell{\emph{query}/\emph{key} dim}
	    & 1 & 2 & 4 & 8 & 16 \\
		\hline 
        LR-Net-26 & 75.7 & 75.4 & 75.1 & 74.7 & 73.7 \\
		\Xhline{2\arrayrulewidth}
	\end{tabular}
	\label{table:key_dim}
\end{table}

\begin{table}[t]
	\setlength{\tabcolsep}{4pt}
	\centering
    \caption{Ablation on channel sharing (top-1 acc \%).}
    \begin{tabular}{c|c|c|c|c|c|c}
		\Xhline{2\arrayrulewidth}
		\makecell{chn. sharing $m$} &
	    1 & 2 & 4 & 8 & 16 & \#chn. \\
	    \hline
        LR-Net-26 & 75.3 & 75.5 & 75.5 & 75.7 & 75.3 & 70.9 \\
		\Xhline{2\arrayrulewidth}
	\end{tabular}
	\label{table:group_num}
\end{table}

\begin{table}[t]
	\setlength{\tabcolsep}{6pt}
	\centering
    \caption{Ablation on appearance composability term and the normalization method (top-1 acc \%).}
    \begin{tabular}{c|c|c|c||c|c}
		\Xhline{2\arrayrulewidth}
	    \multirow{2}{*}{method} & \multicolumn{3}{c||}{app. comp. Eqn.} & \multicolumn{2}{c}{normalization} \\
	    \cline{2-6}
	    & (\ref{eq.square_diff}) & (\ref{eq.abs_diff}) & (\ref{eq.multiplication}) & none & softmax \\
		\hline 
        LR-Net-26 & 75.7 & 75.5 & 75.7 & 74.8 & 75.7 \\
		\Xhline{2\arrayrulewidth}
	\end{tabular}
	\label{table:app_term}
\end{table}

\begin{table}[t]
	\setlength{\tabcolsep}{6pt}
	\centering
    \caption{Comparison with non-local neural networks.}
    \begin{tabular}{c|c|c|c|c}
		\Xhline{2\arrayrulewidth}
	    method & top-1 & top-5 & \# params & FLOPs\\
	    \hline
	    ResNet-26 & 72.8 & 91.0 & 16.0M & 2.6G \\
	    \hline
	    NL-26 & 47.7 & 72.1 & 17.3M & 2.6G \\
	    ResNet-26-NL & 73.4 & 91.2 & 38.2M & 5.6G \\
	    \hline
	    LR-Net-26 & 75.7 & 92.6 & 14.7M & 2.6G \\
	    LR-Net-26-NL & 76.0 & 92.8 & 37.1M & 5.6G \\
		\Xhline{2\arrayrulewidth}
	\end{tabular}
	\label{table:nonlocal}
\end{table}

\begin{table}[t]
	\setlength{\tabcolsep}{8pt}
	\centering
    \caption{Applied on Different Architectures. For LR-Net-18, $\alpha$ balances increasing \# params and decreasing FLOPs.}
    \begin{tabular}{c|c|c|c|c}
		\Xhline{2\arrayrulewidth}
	    method & top-1 & top-5 & \# params & FLOPs\\
	    \hline
	    ResNet-18 & 70.1 & 89.4 & 11.7M & 3.1G\\
	    LR-Net-18 & 74.6 & 92.0 & 14.4M & 2.5G \\
	    \hline
	    ResNet-50 & 76.3 & 93.2 & 25.5M & 4.3G \\
	    LR-Net-50 &  77.3 & 93.6 & 23.3M & 4.3G \\
	    \hline
	    ResNet-101 & 77.9 & 94.0 & 44.4M & 8.0G \\
	    LR-Net-101 & 78.5 & 94.3 & 42.0M & 8.0G \\
		\Xhline{2\arrayrulewidth}
	\end{tabular}
	\label{table:architecture}
	\vspace{-.5em}
\end{table}

\begin{table}[t]
	\setlength{\tabcolsep}{4pt}
	\centering
    \caption{Comparison of robustness to \emph{white-box} adversarial attacks for different architectures on ImageNet (top-1 acc \%).}
    \begin{tabular}{c|c|c|c|c}
		\Xhline{2\arrayrulewidth}
	    \multirow{2}{*}{network} & \multicolumn{3}{c|}{adversarial train} & regular train\\
	    \cline{2-5}
	    & clean & \makecell{\emph{targeted}} & \makecell{\emph{untargeted}} & clean \\
	    \hline
	    ResNet-26 & 44.9 & 37.9 & 14.4 & 72.8 \\
        ResNet-50 & 52.0 & 43.0 & 22.5 & \textbf{76.3} \\
        \hline
        LR-Net-26 & \textbf{52.1} & \textbf{44.2} & \textbf{26.8} & 75.7 \\
		\Xhline{2\arrayrulewidth}
	\end{tabular}
	\label{table:adversarial}
\end{table}

\vspace{.3em}
\noindent \emph{a) Importance of locality} 
Existing bottom-up methods typically compute spatial aggregation over the entire input feature map~\cite{wang2018non,sabour2017dynamic}. We first compare the proposed local relation networks, which enforces a \emph{locality} constraint on spatial aggregation scope, to the equivalent method without this constraint (the ``full image'' column in Table~\ref{table:locality})\footnote{We follow~\cite{wang2018non} to reduce the computation complexity of the ``full image'' method, by adopting downsampled \emph{key} feature maps at high resolution stages: 4$\times$ for \emph{res2}, 2$\times$ for \emph{res3} and 2$\times$ for \emph{res4}. Without this, the accuracy of ``full image'' methods would be even lower.}.

Without encoding any geometric priors (noted as ``NG'' in the table), we observe a huge improvement by changing the aggregation computation from using the whole input feature map to just a 7$\times$7 neighborhood (from 50.7 to 71.9). Surprisingly, while the effectiveness of convolution networks is ascribed to the explicit modeling of geometric priors, we obtain competitive accuracy on ImageNet classification \emph{purely by applying a locality constraint to a geometric-free aggregation method (71.9 vs. 72.8), demonstrating the effectiveness of the locality constraint.}

For the LR-Net-26 models which encode the geometric prior term described in Section~\ref{sec.local_relation_layer}, we also observe significant accuracy improvement, from 68.4 to 75.7. Noting that geometric priors can also act as a method to limit the aggregation scope (positions with smaller geometric prior values will contribute little to the final aggregation computation), the locality constraint further constrains the aggregation scope. 

The locality constraint may also provide an information bottleneck to the network, which aids representation learning.

\vspace{.3em}
\noindent \emph{b) LR-Net Benefits from large kernel}

The regular ResNet-26 model has similar accuracy with 3$\times$3 and 5$\times$5 kernels and loses accuracy when kernel size is larger than 5$\times$5. For ResNet-DW-26 models, the accuracy is almost unchanged when moving from 3$\times$3 to 9$\times$9.

In contrast, both LR-Net-26 variants (with/without geometric prior terms) obtain steadily improved accuracy when the kernel size grows from 3$\times$3 to 7$\times$7: 70.8$\rightarrow$ 71.5$\rightarrow$71.9 for LR-Net-26 (NG) which has no geometric prior term, and 73.6 $\rightarrow$ 74.9 $\rightarrow$ 75.7 for LR-Net-26 which includes both the appearance composability and geometric prior terms. The results demonstrate the superiority of the proposed LR-Net in harnessing large kernels.

\paragraph{Effect of geometric prior} In the last three rows of Table~\ref{table:locality}, encoding of geometric priors is ablated. Both geometric prior embedding methods perform better than that without geometric priors for all spatial scopes, demonstrating their usefulness in visual feature learning.

Comparing the two geometric prior encoding methods, applying a small network on relative positions (the last row) performs better than directly learning independent geometric prior values. The gap between them is larger when the kernel size is larger (0.4 at 3$\times$3 and 3.1 at 9$\times$9), showing that it is crucial to additionally account for relative positions, especially when the neighborhood is large.

\begin{figure}
  \centering
  \includegraphics[width=\linewidth]{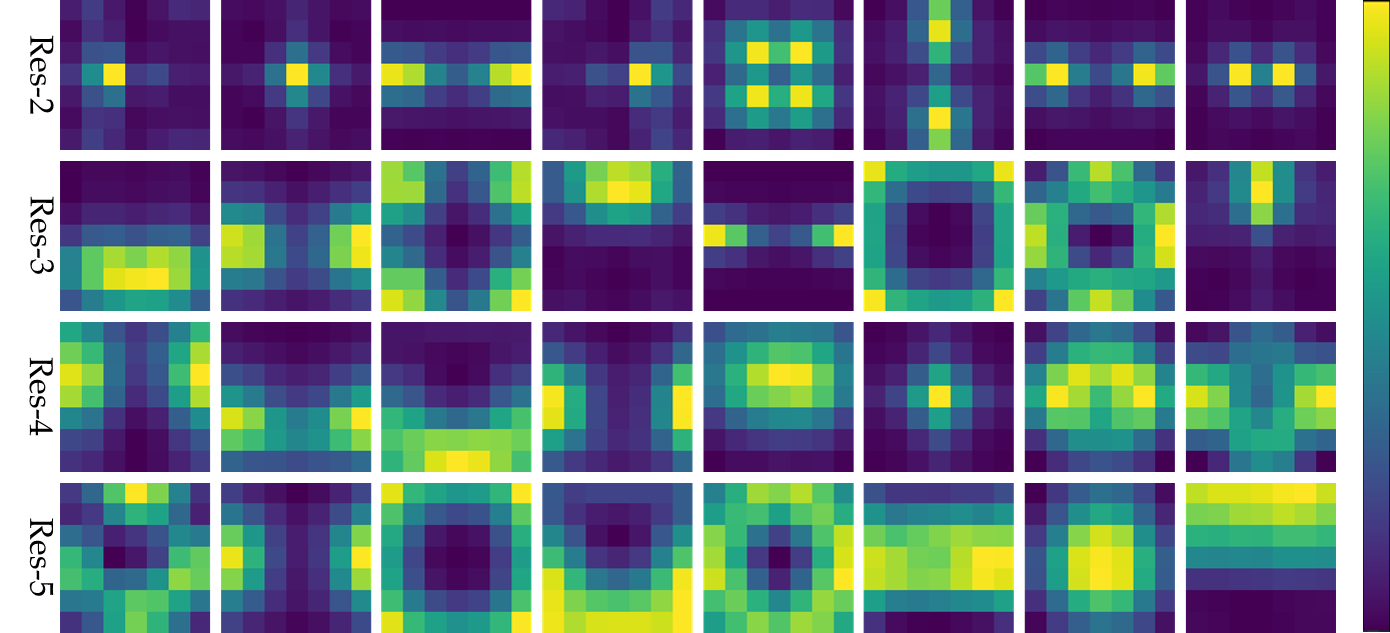}
\caption{Illustration of learnt geometric prior values.}
\label{fig.geometric}
\vspace{-.5em}
\end{figure}

Figure~\ref{fig.geometric} shows the learnt 7$\times$7 geometric prior values after softmax at four stages of LR-Net-26. In general, for lower layers, the priors are sharper, indicating preference for stronger constraints in the learning of appearance composability. For higher layers, the priors are smoother, indicating preference for greater freedom.

\paragraph{Other designs} We also ablate various design elements.

\vspace{.3em}
\noindent \emph{a) Effect of query/key dim}

Table~\ref{table:key_dim} ablates the accuracy of the proposed LR-Net-26 model with varying \emph{key}/\emph{query} dimensions. We follow~\cite{vaswani2017attention} to compute the appearance composablity between \emph{key} and \emph{query} vectors. We find decreased accuracy with increasing \emph{key}/\emph{query} dimension, indicating the superiority of \emph{scalars} over typically-used vectors, as well as a better speed-accuracy tradeoff. 

\vspace{.3em}
\noindent \emph{b) Effect of channel sharing}

Table~\ref{table:group_num} ablates the LR-Net-26 model with varying numbers of shared channels per aggregation ($m$). The accuracy of LR-Net-26 is maintained when $m$ is as large as 8, while being 3$\times$ faster than not sharing ($m=1$).

\vspace{.3em}
\noindent \emph{c) Composability term}

\begin{figure}
  \centering
  \includegraphics[width=0.95\linewidth]{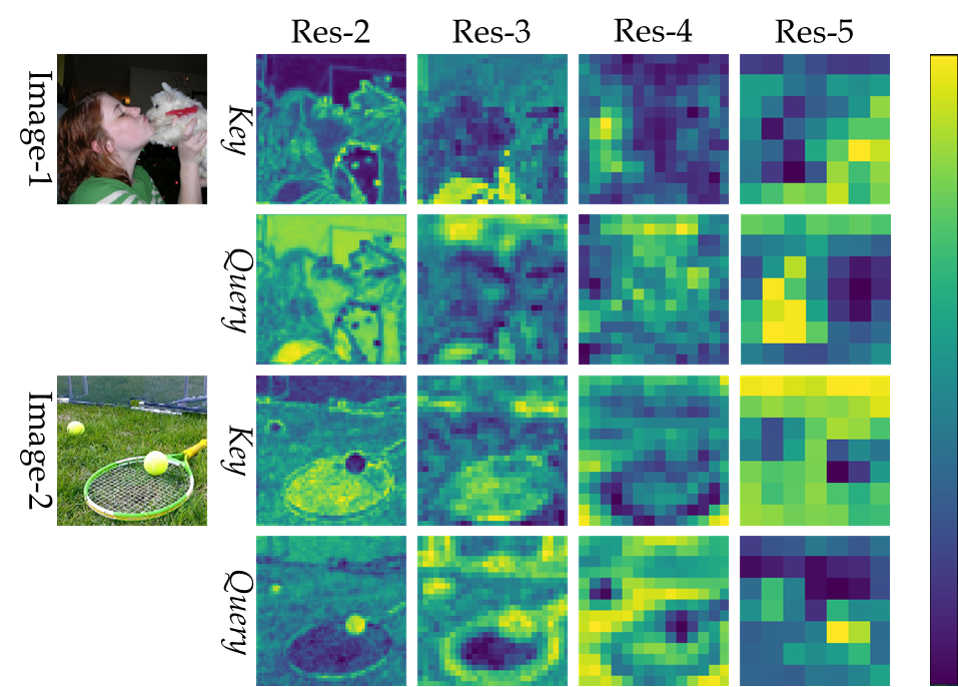}
\caption{Illustration of learnt \emph{key} and \emph{query}.}
\label{fig.key_query}
\vspace{-1.em}
\end{figure}

Table~\ref{table:app_term} ablates over different appearance composabilty terms: Eqn.~(\ref{eq.square_diff}), Eqn.~(\ref{eq.abs_diff}) and Eqn.~(\ref{eq.multiplication}). They are found to work comparably well. Figure~\ref{fig.key_query} exhibits representative examples of \emph{key} and \emph{query} maps learnt using the default term of Eqn.~(\ref{eq.square_diff}), which indicate that composability between semantic visual elements are learnt (girl and dog, tennis ball and racket).

\vspace{.3em}
\noindent \emph{d) Softmax normalization}

Table~\ref{table:app_term} shows that including the softmax normalization in Eqn.~(\ref{eq.local_relation_aggregation_weight}) improves accuracy by 0.9, indicating the importance of normalization in balancing the two terms.

\vspace{-.3em}
\paragraph{Comparison with other bottom-up methods} 
Table~\ref{table:nonlocal} compares LR-Net with other bottom-up methods, i.e. non-local neural networks~\cite{wang2018non}. By directly replacing the 3$\times$3 convolution layer in the ResNet-26 model by non-local modules, the model (NL-26) achieves an accuracy of 47.7, far lower than its regular counterpart. By applying the non-local modules after every residual block, top-1 accuracy of 73.4 is obtained, which is 0.6 higher than its regular counterpart, with about 2$\times$ more computation.

The local relation layer is designed to replace convolution layers for better representation power. It achieves a 2.9 gain over the regular ResNet counterpart with a similar computation load. We note that the non-local module is complementary to local relation networks, bringing a 0.3 gain when applied after every local relation block (see the last row).

\vspace{-.3em}
\paragraph{On different/deeper networks} In Table~\ref{table:architecture}, we evaluate LR-Net with different/deeper network architectures, including ResNet-18 which consists of 8 \emph{basic} residual blocks and ResNet-50/101 which use the same type of \emph{bottleneck} residual blocks but have more layers (50 and 101 layers). The proposed networks are also effective on these architectures.

\subsection{Robustness to adversarial attacks}

We test the ability of LR-Net to withstand adversarial attacks using the \emph{white-box} multi-step PGD attack method~\cite{madry2017towards,kannan2018adversarial}, under both \emph{targeted} and \emph{untargeted} attacks. \emph{Targeted} attacks randomly choose one wrong class as the target, while \emph{untargeted} attacks succeed as long as the model produces wrong predictions. We utilize the hyper-parameters from~\cite{kannan2018adversarial} of the attacking methods, and employ the \emph{targeted} multi-step PGD adversarial method for training with the same hyper-parameters except for the number of attack steps, set to 16 due to limited GPU resources.

Table~\ref{table:adversarial} compares the robustness of LR-Net-26 and the regular ResNet-26/ResNet-50 models against \emph{white-box} adversarial attacks on ImageNet. The LR-Net-26 model performs significantly better than ResNet-26 model against both the \emph{targeted} (+6.3) and \emph{untargeted} attacks (+12.4). The LR-Net-26 model also performs better than the ResNet-50 model (+0.8 for \emph{targeted} attacks and +4.3 for \emph{untargeted} attacks), which uses about 2$\times$ more FLOPs and has better top-1 accuracy in regular training (see the last column of Table~\ref{table:adversarial}). These results indicate that the superior performance of LR-Net in adversarial robustness is not purely due to larger capacity but also because of the architecture itself.

\section{Conclusion and Future Works}

This paper presents the local relation layer, a basic image feature extractor following the general philosophy of introducing compositionality into representation. A deep network composed by this new layer demonstrates strong results on ImageNet classification, significantly expanding the practicality of bottom-up methods, which are long believed to be more fundamental in representation than top-down methods such as convolution.

We note that the study of this new layer is still at an early stage. Future directions include: 1) better GPU memory scheduling for faster implementation; 2) better designs to outperform advanced convolution methods such as deformable convolution~\cite{dai2017deformable,zhu2018deformable}; 3) exploring other properties and the applicability on other vision tasks.

\appendix

\renewcommand{\thesection}{A\arabic{section}}  

\section{Implementation Details}
All architectures take a 3$\times$224$\times$224 image as input. The architectures use a skip connection for the shortcut branch of all residual blocks except for across stages where a channel transformation layer followed by batch normalization is used. In \emph{res3}, \emph{res4} and \emph{res5}, downsampling is applied on the $3\times 3$ convolution layer or the local relation layer in the first residual blocks. For fair comparison in ablation experiments, we adapt the bottleneck ratio $\alpha$ to ensure the same FLOPs for different architectures.

In training, the randomly cropped images and employ scale and aspect ratio augmentation. We perform SGD optimization with a mini-batch of 1024 on 16 GPUs for all experiments except for the experiments of adversarial training in which 32 GPUs are used. The initial learning rate is 0.4, with linear warm-up in the first 5 epochs, and decays by 10$\times$ at the 30th, 60th and 90th epochs, respectively, following~\cite{goyal2017accurate}. The total learning period is 110 epochs, with a weight decay of 0.0001 and momentum of 0.9. In inference, we use a single 224$\times$224 center crop from the resized images with a shorter size of 256. Top-1 and top-5 accuracy are reported.

{\small
\bibliographystyle{ieee}
\bibliography{egbib}
}

\end{document}